\documentclass[10pt,twocolumn,letterpaper]{article}

\usepackage{cvpr}
\usepackage{times}
\usepackage{epsfig}
\usepackage{graphicx}
\usepackage{amsmath}
\usepackage{amssymb}
\usepackage{grffile}
\usepackage{subfigure}
\usepackage{multirow}
\usepackage{booktabs}
\usepackage{enumitem}
\usepackage{mdwlist} 
\usepackage{bm}
\usepackage{color,xcolor}
\definecolor{orange}{rgb}{1,0.4,0} 

\usepackage[breaklinks=true,bookmarks=false]{hyperref}

\cvprfinalcopy 

\def\cvprPaperID{6548} 

\ifcvprfinal\fi
\begin{document}
    
    \title{Single Image Reflection Removal through Cascaded Refinement} 
    \author{Chao Li$^{1, *}$  \quad Yixiao Yang$^{1,}$\thanks{The first two authors contribute equally.} \quad  Kun He$^{1,}$\thanks{Corresponding author.}  
    \quad  Stephen Lin$^2$ \quad John E. Hopcroft$^3$ \\
    \small
    $^1$School of Computer Science and Technology, Huazhong University of Science and Technology\\
    \small
    $^2$Microsoft Research Asia,  
    \small
    $^3$Computer Science Department, Cornell University\\
    {\tt\small brooklet60@hust.edu.cn}}
    \maketitle
    \thispagestyle{empty}
    
    
\begin{abstract}
We address the problem of removing undesirable reflections from a single image captured through a glass surface, which is an ill-posed, challenging but practically important problem for photo enhancement. Inspired by iterative structure reduction for hidden community detection in social networks, we propose an Iterative Boost Convolutional LSTM Network (IBCLN) that enables cascaded prediction for reflection removal.
IBCLN is a cascaded network that iteratively refines the estimates of transmission and reflection layers in a manner that they can boost the prediction quality to each other, and information across steps of the cascade is transferred using an LSTM. 
The intuition is that the transmission is the strong, dominant structure while the reflection is the weak, hidden structure. They are complementary to each other in a single image and thus a better estimate and reduction on one side from the original image leads to a more accurate estimate on the other side.
To facilitate training over multiple cascade steps, we employ LSTM to address the vanishing gradient problem, and propose residual reconstruction loss as further training guidance. 
Besides, we create a dataset of real-world images with reflection and ground-truth transmission layers to mitigate the problem of insufficient data. 
Comprehensive experiments demonstrate that the proposed method can effectively remove reflections in real and synthetic images compared with state-of-the-art reflection removal methods.
\end{abstract}
\section{Introduction}
Undesirable reflections from glass occur frequently in real-world photos. It not only significantly degrades the image quality, but also affects the performance of downstream computer vision tasks like object detection and semantic segmentation. As the reflection removal problem is ill-posed, early works primarily tackle it with multiple input images~\cite{szeliski2000layer, sarel2004separating, li2013exploiting, xue2015computational, guo2014robust,  sun2016automatic,  gai2011blind, han2017reflection}. More recently, researchers attempt to address the more common and practically significant scenario of a single input image~\cite{levin2003learning, levin2004separating, li2013exploiting, li2014single, wan2016depth, springer2017reflection, arvanitopoulos2017single, tuncer2011ground}. 

For single-image reflection removal (SIRR), researchers have observed that some handcrafted priors may help for distinguishing the transmission layer from the reflection layer in a single image. But these priors often do not generalize well to different types of reflections and scenes owing to disparate imaging conditions. In recent years, researchers apply data-driven learning to replace handcrafted priors via deep convolutional neural networks. With abundant labeled data, a network can be trained to perform effectively over a broad range of scenes. 
However, learning-based single-image methods still have much room for improvement due to complications such as limited training data, disparate imaging conditions, varying scene content, limited physical understanding of this problem, and the performance limitation of various models. 

\begin{figure*}[htbp]
    \centering
    \includegraphics[width=1\textwidth]{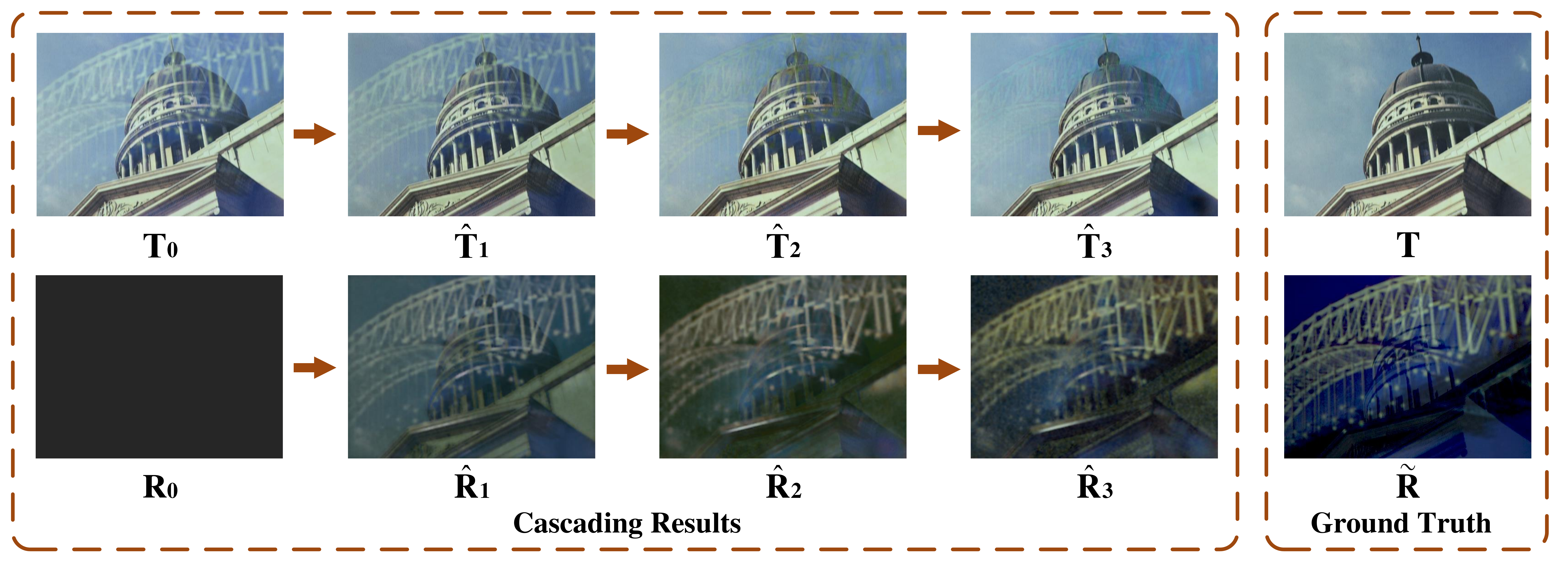}
    \caption{Visualization of results at different cascade steps of the two sub-networks in the proposed model. The estimates of transmissions and residual reflections become increasingly more accurate as they progress through the cascade. More results are in the \textit{suppl. material}.
    }
    \vspace{-0.2em}
    \label{fig:cascading-results}
\end{figure*}

In this work, inspired by the iterative structure reduction approach for hidden community detection in social networks~\cite{he2018hidden, He15corr}, we introduce a cascaded neural network model for transmission and reflection decomposition. Figure \ref{fig:cascading-results} illustrates the cascade results in our model, where the transmission and reflection are progressively refined during the iterations. 
To the best of our knowledge, previous works on reflection removal did not utilize a cascaded refinement approach. 
Though some methods such as BDN~\cite{yang2018seeing} obtain predictions over a sequence of a few sub-networks, they do not iteratively refine the estimates, but rather they conduct a short alternating optimization, e.g., by estimating the reflection from the input image and the initial transmission layer, and then estimating the transmission from the input image and the estimated reflection layer. 

For a cascade model on SIRR, a simple approach is to employ one network to generate a predicted transmission that serves as the auxiliary information of the next network, and continue such process with subsequent networks to iteratively improve the prediction quality. 
With a long cascade, however, the training becomes difficult due to the vanishing gradient problem and limited training guidance at each step.
To address this issue, we design a convolutional LSTM (Long Short-Term Memory) network, which saves information from the previous iteration (i.e. time step) and allows gradients to flow unchanged.

In our model, two sub-networks use identical convolutional LSTM architecture, one for transmission prediction and the other for reflection prediction. They share input information using the outputs of the previous time step to boost each other’s effectiveness. Here we propose a residual reconstruction loss as further training supervision at each cascade step. To simplify the reconstruction loss, we define a new concept of residual reflection, which will be described in Sec.~\ref{sec:loss}.

Though a few real-world datasets with ground-truth have been presented~\cite{wan2017benchmarking, zhang2018single}, the real-world data for SIRR is still insufficient due to the tremendously labor-intensive work.  
To help resolve the insufficiency of the real-world training data, we also collect a real dataset with densely-labeled ground truth in disparate imaging conditions and varying scenes.

Our main contributions are as follows:
\begin{itemize}[noitemsep,topsep=0pt]
    \setlength{\itemsep}{1pt}
    \setlength{\parskip}{0pt}
    \setlength{\parsep}{0pt}
    \item 
     We propose a new network architecture, a cascaded network, with loss components that achieves state-of-the-art quantitative results on real-world benchmarks for the single image reflection removal problem.
    \item We design a residual reconstruction loss, which can form a closed loop with the linear method for synthesizing images with reflections, to expand the influence of the synthesis method across the whole network.
    \item We collect a new real-world dataset containing images with densely-labeled ground-truth, which can serve as baseline data in future research. 
\end{itemize}

\section{Related Work}
Mathematically speaking, SIRR operates on a captured image $\mathbf{I}$, which is generally assumed to be a linear combination of a transmission layer $\mathbf{T}$ and a reflection layer $\mathbf{R}$. 
The goal is to infer a transmission layer $\mathbf{T}$ that is free of reflections. 
In this work, we focus on deep learning-based SIRR, which has produced state-of-the-art results. Previous multiple-image methods~\cite{xue2015computational, guo2014robust, li2013exploiting, sun2016automatic, sarel2004separating, gai2011blind, szeliski2000layer, han2017reflection} and single-image-priors based methods~\cite{levin2004separating, li2014single, levin2003learning, springer2017reflection, arvanitopoulos2017single, wan2016depth, li2013exploiting, tuncer2011ground} are not considered here.

Due to the advantages in robustness and performance, there is an emerging interest in applying neural networks to SIRR. Fan \etal~\cite{fan2017generic} provide the first neural network model to solve this ill-posed problem. They propose a linear method for synthesizing images with reflection for training, and use an edge map as auxiliary information to guide the reflection removal. Wan \etal~\cite{wan2018crrn} develop two cooperative sub-networks, which predict the transmission layer intensity and gradients concurrently. Both of these works~\cite{fan2017generic, wan2018crrn} utilize edge or gradient information of the captured layer $\mathbf{I}$, motivated by the idea that the reflection layers are usually not in focus and thus blurry as compared to the transmission layers. 
From the edge information of the captured image $\mathbf{I}$, the edge map of the transmission image $\mathbf{T}$ is predicted and used in estimating the transmission result.
Instead, BDN~\cite{yang2018seeing} predicts reflection layers which are then used as auxiliary information in a subsequent network to estimate the transmission.

In several recent methods, improved formulations of the objective function are presented. These include the adoption of perceptual losses~\cite{johnson2016perceptual} to account for both low-level and high-level image information~\cite{chi2018single, jin2018learning, zhang2018single}. In these works, images are fed to a deep network pre-trained on ImageNet, and comparisons are made based on extracted multi-stage features. Adversarial losses have also been applied, specifically to improve the realism of predicted transmission layers~\cite{zhang2018single, lee2018generative, wen2019single, wei2019single}.

Another direction of study focuses on datasets for training. Moving beyond improvements for the linear synthesis method in~\cite{fan2017generic} and~\cite{zhang2018single}, Wen \etal~\cite{wen2019single} synthesize training data with learned non-linear alpha blending masks that better model the real-world imaging conditions. These masks are also used in forming a reconstruction loss that guides the prediction of transmission layers. To deal with the insufficiency of densely-labeled training data, Wei \etal~\cite{wei2019single} present a technique for utilizing misaligned real-world images as the training data, as they are less burdensome to acquire than aligned images and are more realistic than synthetic images.
\begin{figure*}[htbp]
    \begin{center}
    \includegraphics[width=0.9\textwidth]{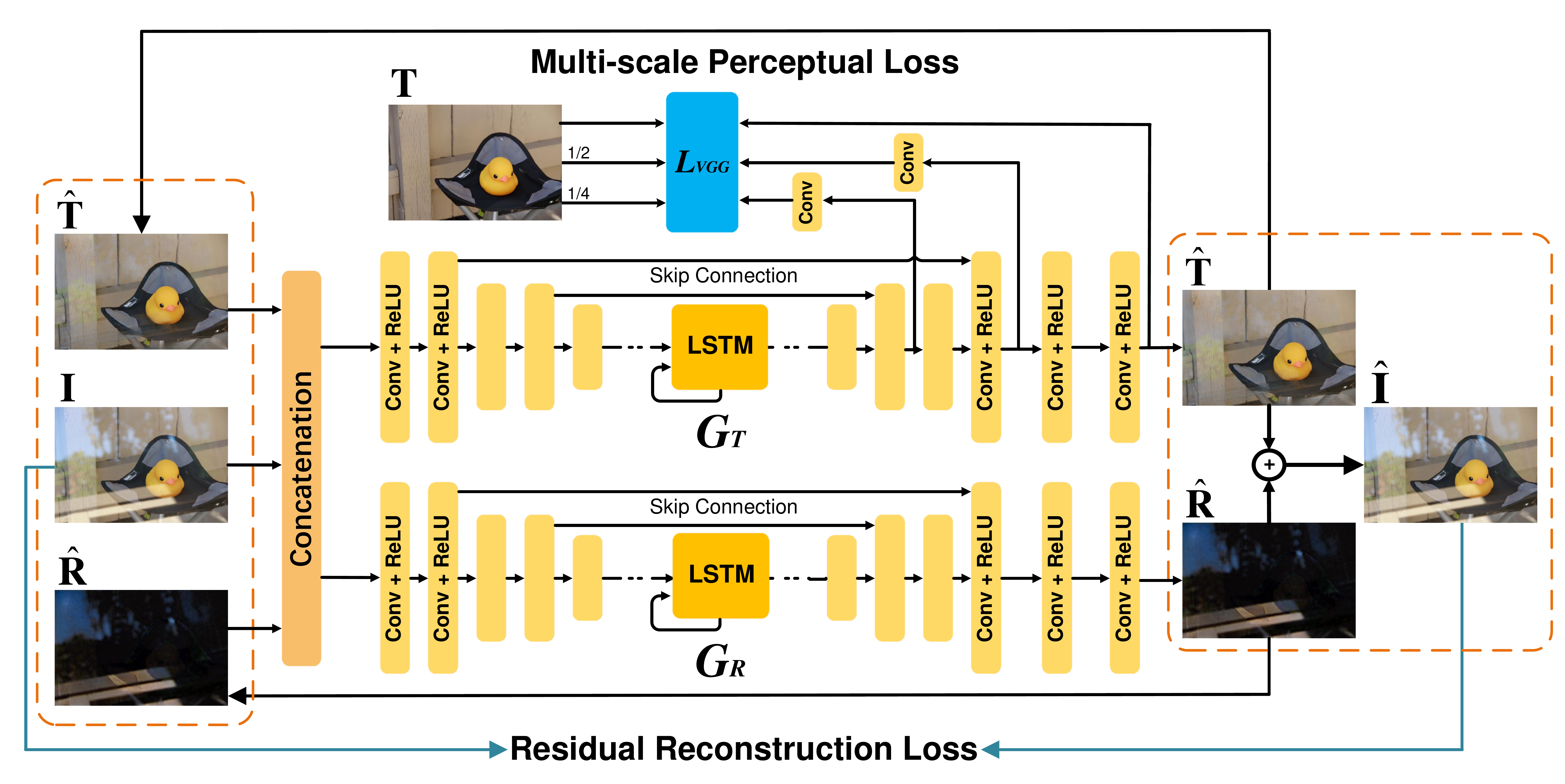}
    \end{center}
    \vspace{-0.5em}
    \caption{The architecture of IBCLN. The cascaded network consists of a transmission generative sub-network $\bm{G_T}$ and a reflection generative sub-network $\bm{G_R}$ with skip connections, both of which are convolutional LSTM networks. The images generated at each time step by the two sub-networks will be fed back at the next time step. The overall network is trained in an end-to-end manner.
    }
    \vspace{-0.5em}
    \label{fig:network}
\end{figure*}

\section{Proposed Method}

\subsection{Motivation} \label{sec:motivation}

This work is motivated by research on hidden structures in social networks.
He~\etal~\cite{he2018hidden,He15corr} define a set of communities as hidden structure if most of the members also belong to other stronger communities. They propose an iterative boost approach to separate a set of strong, dominant communities and another set of weak, hidden communities, and boost the detection accuracy on both sides. The key idea is that, when they detect an approximate set of dominant communities using a base algorithm, and weaken their internal connection to the average connection of the overall graph, the dominant structure is reduced to boost the detection on the set of hidden communities, and vice versa.

Under the scenario of SIRR, a useful trick is to employ sub-networks to learn auxiliary information that can facilitate transmission layer prediction. The types of auxiliary information utilized in existing works include edge information~\cite{fan2017generic, wan2018crrn} and predicted reflections~\cite{yang2018seeing}.
The ideal auxiliary information would be the ground truth reflection-free version of the transmission layer, which is what we seek to predict. As this is not available at inference time, we instead use approximations to the ground-truth transmission in the form of predicted transmissions as the auxiliary information. Though certainly not as useful as the ground truth, it nevertheless provides strong guidance, especially as the transmission predictions improve. The key issue then becomes how to drive the transmission estimations closer and closer to the ground truth.  Referring to the work of He~\etal~\cite{he2018hidden,He15corr}, we regard the transmission layer as the strong, dominant structure, and the reflection layer as the weak, hidden structure. By iteratively reducing the more accurate version of the counterpart, we could extract more accurate approximations on the two layers of images.

Our model contains two sub-networks that can collaborate and boost each other's output by reducing the output of one side from the original image as effective auxiliary information for the other complementary side. Such collaborative cascaded refinement of the dominant image (transmission) and the weak image (reflection) is novel for the training of a neural network.  

\subsection{General Design Principles} \label{sec:design}
We use two convolutional LSTM networks to separately generate the predicted transmission layers and the predicted reflection layers. The input of each sub-network includes the outputs of both the transmission and reflection sub-networks. Besides, the outputs of the two sub-networks are combined within a reconstruction loss to supervise the whole model at each time step. The synergy between the two sub-networks leads to a mutual boost in their predictions, resulting in progressive improvements of the auxiliary information and finally accurate estimates of the transmission.

To ensure that the transmission sub-network and the reflection sub-network generate complementary outputs, we enforce a reconstruction loss where the image $\mathbf{\hat{I}}$ synthesized from the estimated transmission and reflection is expected to match the input image $\mathbf{I}$. 

A related constraint is employed in RmNet~\cite{wen2019single}, which synthesizes an image $\mathbf{I}$ from the 
ground-truth transmission layer with no reflection,
the reflection layer used to produce reflections off the glass, and an alpha blending mask $\mathbf{W}$. Thus, 
$\mathbf{I} = \mathbf{W} \circ \mathbf{T} + (\mathbf{1} - \mathbf{W}) \circ \mathbf{R}$,  
where $\circ$ denotes element-wise multiplication. 
The reconstructed image $\mathbf{\hat{I}}$ is then compared to the synthetic input image $\mathbf{I}$. However, their alpha blending model only approximates the complex physical mechanisms involved in forming an actual input image with reflections, as it does not model effects such as spatially varying blurs and Gamma correction~\cite{BULL201499}, which is used to correct for the differences between the way a camera captures content and the way our visual system processes light. This will limit reconstruction quality on real-world input images and consequently degrade prediction results as we found from experiments reported in Table \ref{tb:method-com}. 

To avoid the problem that RmNet encounters, we use a scale parameter $\alpha$ instead of the element-wise mask matrix $\mathbf{W}$, and we directly calculate the {\em residual reflection} $\mathbf{\widetilde{R}}$ by $\mathbf{I} - \alpha \cdot \mathbf{T}$. In this way, we do not require modeling the complicated physical process involved in the formation of images with reflection, and our performance does not suffer from deficiencies in such a synthesis model. The benefit of predicting residual reflection instead of the reflection layer used to produce reflections off the glass is that image reconstruction becomes simplified as just the sum of the predicted transmission and the predicted residual reflection.
Also, different from RmNet, all our linear operations are done in the linear color space, removing Gamma correction~\cite{BULL201499}.

\subsection{Network Architecture}
The architecture of the proposed network is illustrated in Figure \ref{fig:network}\footnote{Code and model: https://github.com/JHL-HUST/IBCLN/.}. IBCLN consists of two sub-networks: a transmission-prediction network $\bm{G_T}$ and a reflection-prediction network $\bm{G_R}$. The two sub-networks are both convolutional LSTM networks with the same architecture but different goals. The former aims to learn the transmission $\mathbf{T}$ while the latter aims to learn the residual reflection $\mathbf{\widetilde{R}}$, so they learn completely different weight parameters. Each sub-network consists of an encoder with 11 Conv-relu blocks that extract the features from the input image, a convolutional LSTM unit \cite{xingjian2015convolutional} and a decoder with 8 convolutional layers for generating the predicted transmission layer or the predicted residual reflection layer. Each convolutional layer is followed by a ReLU activation, except for the LSTM layers which are followed by a Sigmoid activation or a Tanh activation. In each sub-network, there are two skip connections between the encoder and the decoder to prevent blurred outputs. The convolutional layers and skip connections are similar to those of a contextual autoencoder~\cite{qian2018attentive}. Different from previous works, our objective function includes the proposed residual reconstruction loss and a multi-scale perceptual loss. 

\begin{figure}[htbp]
    \begin{center}
        \includegraphics[width=0.47\textwidth]{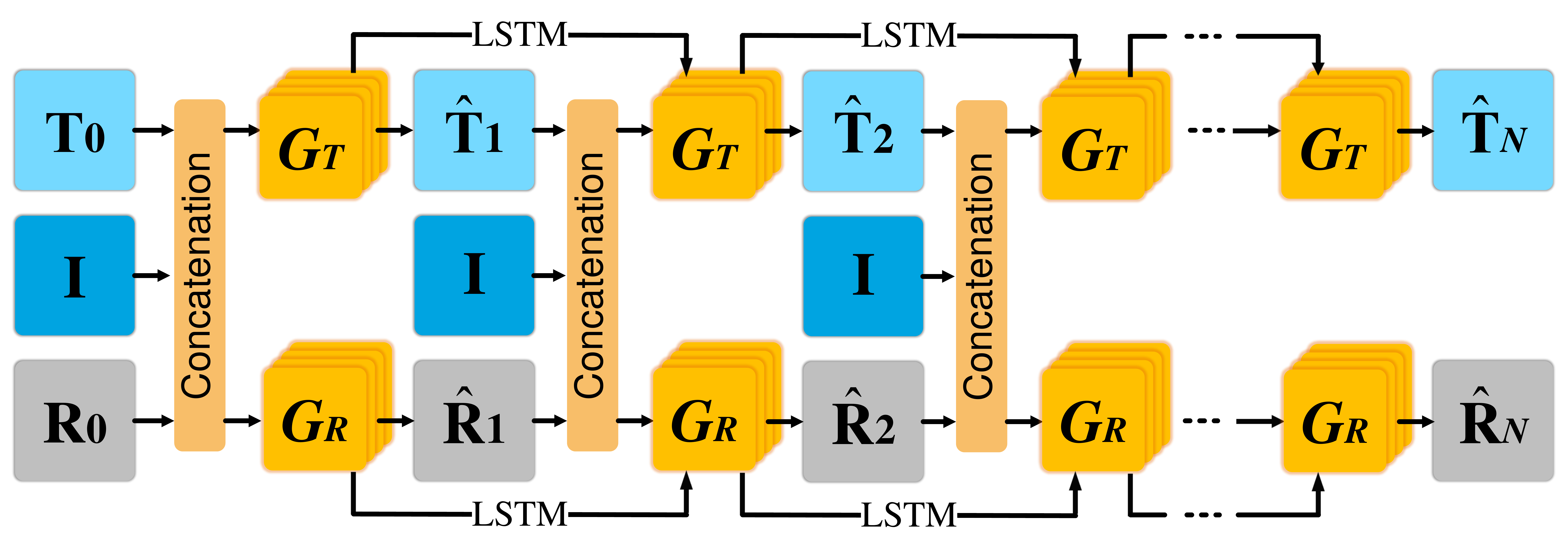}
    \end{center}
    \vspace{-0.5em}
    \caption{Characterizing IBCLN with increasing number of time steps. All blocks labeled as $\bm{G_T}$ indicate one sub-network and all blocks labeled as $\bm{G_R}$ indicate another sub-network. The output at time step $t-1$ serves as the input at time step $t$.  $\mathbf{\hat{T}}_1$, $\mathbf{\hat{T}}_2$, ..., $\mathbf{\hat{T}}_N$ are the predicted transmission. $\mathbf{\hat{R}}_1$, $\mathbf{\hat{R}}_2$, ..., $\mathbf{\hat{R}}_N$ are the predicted residual reflection.
    }
    \label{fig:main-part}
    \vspace{-0.5em}
\end{figure}

Figure \ref{fig:main-part} illustrates IBCLN from a different perspective. All $\bm{G_T}$ illustrated in this figure is exactly the same network with the same parameters, but at different time steps in the cascade. We connect $\bm{G_T}$ at adjacent time steps with convolutional LSTM units that save information from the previous time step. In the actual model, the convolutional LSTM unit is in the middle of the sub-network and connected with convolutional layers. The convolutional LSTM unit has four gates, including an input gate, a forget gate, an output gate, as well as a cell state. The cell state encodes the state information that will be fed to the next LSTM. The LSTM’s output feature is fed into the next convolutional layer. More details can be found in ConvLSTM~\cite{xingjian2015convolutional}. At time step $t$, both of the sub-networks take nine channels of input, specifically a concatenation of the synthetic image $\mathbf{I}$, the predicted transmission $\mathbf{\hat{T}}_{t-1}$ and residual reflection $\mathbf{\hat{R}}_{t-1}$ predicted at time step $t-1$ ($1 < {t} \leq N$). $\mathbf{T}_0$ is set to be the synthetic image $\mathbf{I}$ and $\mathbf{R}_0$ is set to 0.1 for all entries. The output of the transmission prediction network $\bm{G_T}$ at the final time step $N$ serves as the final result.

Many previous works consider auxiliary information to be important for predicting reflection-free transmission layers~\cite{fan2017generic, yang2018seeing, wan2018crrn, wen2019single}, since it indicates to the network where the removal should be focused on. In our work, $\mathbf{\hat{T}}_{t-1}$ and $\mathbf{\hat{R}}_{t-1}$ are saved to serve as the auxiliary information of step $t$ ($1 < {t} \leq N$). The auxiliary information will improve with increasing numbers of time steps (see Figure \ref{fig:cascading-results}). Since the predicted transmissions represent what the network can infer at a given time step,
using them as auxiliary information is effective. 
Additionally, the predicted residual reflection is complementary to the predicted transmission in an image, so it also contains meaningful information. 

Considering that the iterative process may require a long cascade, using conventional convolutional networks as the sub-networks would make the full model hard to train. This motivates our use of two convolutional LSTM networks, each with a convolutional LSTM unit. 
The continuity among time steps makes the model easy to train. 
Additionally, a cascaded architecture has fewer parameters to learn, as both of the sub-networks are iterated multiple times and each instance of a sub-network shares the same weights.
Moreover, a convolutional LSTM network has more complete information exchange either within itself or between the two sub-networks, which is more in line with our iterative boost idea.

\vspace{-0.2em}
\subsection{Objective Function} \label{sec:loss}
\noindent \textbf{Residual Reconstruction Loss.~}
For the existing linear models~\cite{fan2017generic, zhang2018single} for generating synthetic images, the general steps are to perform a series of complex operations on a reflection image to produce a reflection layer $\mathbf{R}$, then to generate a synthetic image $\mathbf{I}$ by a linear operation: $\mathbf{I} = \text{clip} (\alpha \cdot \mathbf{T} + \mathbf{R})$. 
Usually $\alpha \in [0.8, 1]$ due to the slight attenuation of light as it passes through a glass plane.  
The weight of the reflection layer $\mathbf{R}$ is 1 as the original reflection image has been subtracted adaptively by the synthesis method. 
The clipping operation forces all values of the synthetic image to be in [0,1].

We introduce a new loss to the proposed network, called the \textit{residual reconstruction loss}. We adopt the above synthesis model, but replace $\mathbf{R}$ with $\mathbf{\widetilde{R}}$, where $\mathbf{\widetilde{R}}$ is determined from $\mathbf{I}$ and $\mathbf{T}$. $\mathbf{\widetilde{R}}$ offers more effective auxiliary information for transmission prediction, and a more convincing ground truth, as compared to the artificially constructed $\mathbf{R}$. $\mathbf{\widetilde{R}}$ is obtained by reverting the linear synthesis model, as
\begin{equation}
\vspace{-0.2em}
\mathbf{\widetilde{R}}=\mathbf{I}- \alpha \cdot \mathbf{T}.
\vspace{-0.2em}
\label{eq:R}
\end{equation}

With this definition of $\mathbf{\widetilde{R}}$, the clipping operation is not needed and we avoid its loss of information.
After $\mathbf{\widetilde{R}}$ is calculated, it can be used as the ground truth of $\bm{G_R}$ to guide the generation of the predicted residual reflection $\mathbf{\hat{R}}$.
Then, we can simply revert Eq. (\ref{eq:R}) in the objective function, as 
\begin{equation}
\vspace{-0.2em}
\mathbf{\hat{I}}=\alpha \cdot \mathbf{\hat{T}} + \mathbf{\hat{R}},
\vspace{-0.2em}
\label{eq:ITR}
\end{equation}
where $\mathbf{\hat{T}}$, $\mathbf{\hat{R}}$ and $\mathbf{\hat{I}}$ are the predicted transmission, predicted residual reflection and the reconstructed image, respectively. $\alpha$ is the same as in the synthesis model. 

Note that all the above linear operations are done in the linear color space, so the Gamma correction~\cite{BULL201499} on each image is removed before inclusion in linear operations.

It is intuitive that the reconstructed image $\mathbf{\hat{I}}$ should be similar to the original input through a well-trained network.
The residual reconstruction loss is defined as:
\begin{equation}
\vspace{-0.2em}
\mathcal{L}_{residual}= \sum_{I\in \mathcal{D}}\sum\limits^{N}_{t = 1} \mathcal{L}_{MSE}(\mathbf{I}, \mathbf{\hat{I}}_t).
\vspace{-0.2em}
\end{equation}
$\mathcal{L}_{MSE}$ indicates the mean squared error. $t$ denotes the time step of the two sub-networks. $N$ represents the final time step when $\mathbf{\hat{T}}$ converges. 

The residual reconstruction loss works well experimentally. 
One potential reason is that the two sub-networks have the same architecture but complementary objectives. With the same architecture, they may be under-trained or over-trained concurrently. The complementary objectives within the residual reconstruction loss can balance the error from the two sub-networks. If both of the two sub-networks are either under-trained or over-trained, the error will be doubled in the residual reconstruction loss.

\noindent \textbf{Multi-scale Perceptual Loss.~}
Multi-scale losses are effective in image decomposition tasks such as raindrop removal~\cite{qian2018attentive}. A multi-scale loss extracts the features from different decoder layers and feeds them into a convolutional layer to form outputs at different resolutions. The outputs are then compared to those of real images by their $\mathcal{L}_{MSE}$ distance. By adopting such a loss in our task, we can capture more contextual information from various scales. We change the $\mathcal{L}_{MSE}$ distance to the perceptual distance between the predicted image and the real image over different scales. This loss thus considers different scales of both low-level and high-level information. 
We define the loss function as:
\begin{equation}
\vspace{-0.2em}
\label{eq_g}
\begin{split}
\mathcal{L}_{MP} =&\sum_{T, T^3, T^5\in \mathcal{D}} (\mathcal{L}_{VGG}(\mathbf{T}, \mathbf{\hat{T}}) +\gamma_3  \mathcal{L}_{VGG}(\mathbf{T^3},\mathbf{\hat{T}^3})\\
             &~~~~~~~~~~~~~~~~+ \gamma_5 \mathcal{L}_{VGG}(\mathbf{T^5}, \mathbf{\hat{T}^5})),
\end{split}
\vspace{-1.2em}
\end{equation}
where $\mathbf{\hat{T}}$, $\mathbf{\hat{T}^3}$, $\mathbf{\hat{T}^5}$ indicate the outputs of the last, 3$^{rd}$ last and 5$^{th}$ last layers at time step $N$, whose sizes are $1$, $\frac{1}{2}$ and $\frac{1}{4}$ of the original size, respectively. $\mathbf{T}$, $\mathbf{T^3}$ and $\mathbf{T^5}$ indicate the ground truth that has the same scale as that of the outputs, respectively. Layers with smaller size are not considered since their information is relatively insignificant. We set $\gamma_3 = 0.8$ and $\gamma_5 = 0.6$. All the images are fed into the VGG19 network~\cite{simonyan2014very}. We compare the outputs of the layers `conv1\_2' and `conv2\_2’ in the VGG19 network.

\noindent \textbf{Pixel Loss.~}
To ensure that the outputs become as close to the ground truth as possible, we utilize the $\mathcal{L}_{MSE}$ loss to measure the pixel-wise distance between them. Our pixel loss is defined as follows: 
\begin{equation}
\vspace{-0.2em}
\mathcal{L}_{pixel}= \sum_{T\in \mathcal{D}} {\sum\limits^{N}_{t = 1} [\mathcal{L}_{MSE}(\mathbf{T}, \mathbf{\hat{T}}_t)+ \mathcal{L}_{MSE}(\mathbf{\widetilde{R}}, \mathbf{\hat{R}}_t)]},
\vspace{-0.2em}
\end{equation}
where $\mathbf{\widetilde{R}}$ is the residual reflection. $\mathbf{\hat{T}}_t$ and $\mathbf{\hat{R}}_t$ are the outputs at time step $t$.

\noindent\textbf{Adversarial Loss.~}  
To improve the realism of the generated transmission layers, we further add an adversarial loss. We define an opponent discriminator network $\bm{D}$. The adversarial loss is defined as (refer to~\cite{zhang2018single} for details):
\begin{equation}
\vspace{-0.2em}
\mathcal{L}_{adv} =\sum_{T\in \mathcal{D}}{-\log \bm{D}( \mathbf{T},\mathbf{\hat{T}}) }. 
\vspace{-0.2em}
\end{equation}  
    
\noindent\textbf{Overall Loss.~}     
Overall, our objective function of IBCLN is defined as:
\begin{align}
L = \lambda_1 \mathcal{L}_{residual} + \lambda_2 \mathcal{L}_{MP}+ \lambda_3 \mathcal{L}_{pixel} + \lambda_4 \mathcal{L}_{adv},
\label{eq:pixelLoss}
\end{align} 
where we empirically set the weights as $\lambda_1 = 2, \lambda_2 = 1, \lambda_3 = 2, \lambda_4 = 0.01$  throughout our experiments.

\begin{table*}[!htbp]
    \centering
        \caption{Quantitative comparison of different methods on three real-world benchmark datasets. The best results are in \textbf{bold} and \textcolor{orange}{orange} color, and the second best results are \underline{underlined} and in \textcolor{blue}{blue} color. 
        `Average' is obtained by averaging the metric scores of all images from all the above real-world datasets.}
        \footnotesize
        \vspace{1em}
        \setlength{\tabcolsep}{4.5mm}{
            \begin{tabular}{lccccccc}
                \toprule
                \multirow{3}{*}{Dataset (size)} & \multirow{3}{*}{Index} & \multicolumn{6}{c}{Methods} \\ \cline{3-8}
                &  & CEILNet-F & Zhang et al. & BDN-F & RmNet & ERRNet-F & IBCLN \\
                &  & ~\cite{fan2017generic}& ~\cite{zhang2018single}& ~\cite{yang2018seeing}& ~\cite{wen2019single}& ~\cite{wei2019single}& \\
                \hline

                \multirow{2}{*}{Object (200)}
                &PSNR&22.81&22.68&23.02&20.33& \textcolor{blue}{\underline{24.85}}&\textcolor{orange}{\textbf{24.87}}\\
                &SSIM&0.801&0.874&0.853&0.793&\textcolor{blue}{\underline{0.889}}&\textcolor{orange}{\textbf{0.893}} \\
                \hline
                
                \multirow{2}{*}{Postcard (199)}
                
                &PSNR&20.08&16.81&20.71&19.71&\textcolor{blue}{\underline{21.99}}&\textcolor{orange}{\textbf{23.39}}\\
                
                &SSIM&0.810&0.797&0.857&0.808&\textcolor{blue}{\underline{0.874}}&\textcolor{orange}{\textbf{0.875}}\\
                \hline
                \multirow{2}{*}{Wild (55)}    
                &PSNR&22.14&21.52&22.34&21.98&\textcolor{blue}{\underline{24.16}}&\textcolor{orange}{\textbf{24.71}}\\    
                &SSIM& 0.819&0.829&0.821&0.821&\textcolor{blue}{\underline{0.847}}&\textcolor{orange}{\textbf{0.886}} \\    
                \hline

                \multirow{2}{*}{Zhang et al. (20)}
                &PSNR&18.79&\textcolor{blue}{\underline{22.42}}&19.47&18.77& \textcolor{orange}{\textbf{23.35}}&21.86\\
                &SSIM&0.749&\textcolor{blue}{\underline{0.792}}&0.720&0.681& \textcolor{orange}{\textbf{0.811}}& 0.762\\
                \hline
                \multirow{2}{*}{Nature (20)}
                &PSNR& 19.33&19.56&18.92&19.36&\textcolor{blue}{\underline{22.18}}& \textcolor{orange}{\textbf{23.57}}  \\
                &SSIM&   0.745 &0.736 &0.737&0.725&\textcolor{blue}{\underline{0.756}}&\textcolor{orange}{\textbf{0.783}}\\
                \hline
                \multirow{2}{*}{\emph{Average} (494)} 
                &PSNR&21.31&20.85&21.68&20.19&\textcolor{blue}{\underline{23.45}}&\textcolor{orange}{\textbf{24.08}}\\
                &SSIM&0.806&0.829&0.841 &0.795&\textcolor{blue}{\underline{0.870}}&\textcolor{orange}{\textbf{0.875}}\\
                \bottomrule
        \end{tabular}}
        \label{tb:method-com}
        \vspace{2pt}
    \end{table*}

\subsection{Implementation Details}
We implement the proposed IBCLN in Pytorch on a PC with an Nvidia Geforce GTX 2080 Ti GPU. The overall model is trained for 80 epochs with a batch size of 2, using the Adam optimizer~\cite{kingma2014adam}. The learning rate for the overall network training is set to 0.0002. For the training data, we use 4000 images containing 2800 synthetic images and 1200 image patches of size $256 \times 256$ from 290 real-world training images, containing 200 images from our created dataset and 90 images from Zhang et al.~\cite{zhang2018single}.

\section{Experiments}
\subsection{Dataset Preparation}
\begin{figure}[h]
    \vspace{-1em}
    \begin{center}
        \includegraphics[width=0.47\textwidth]{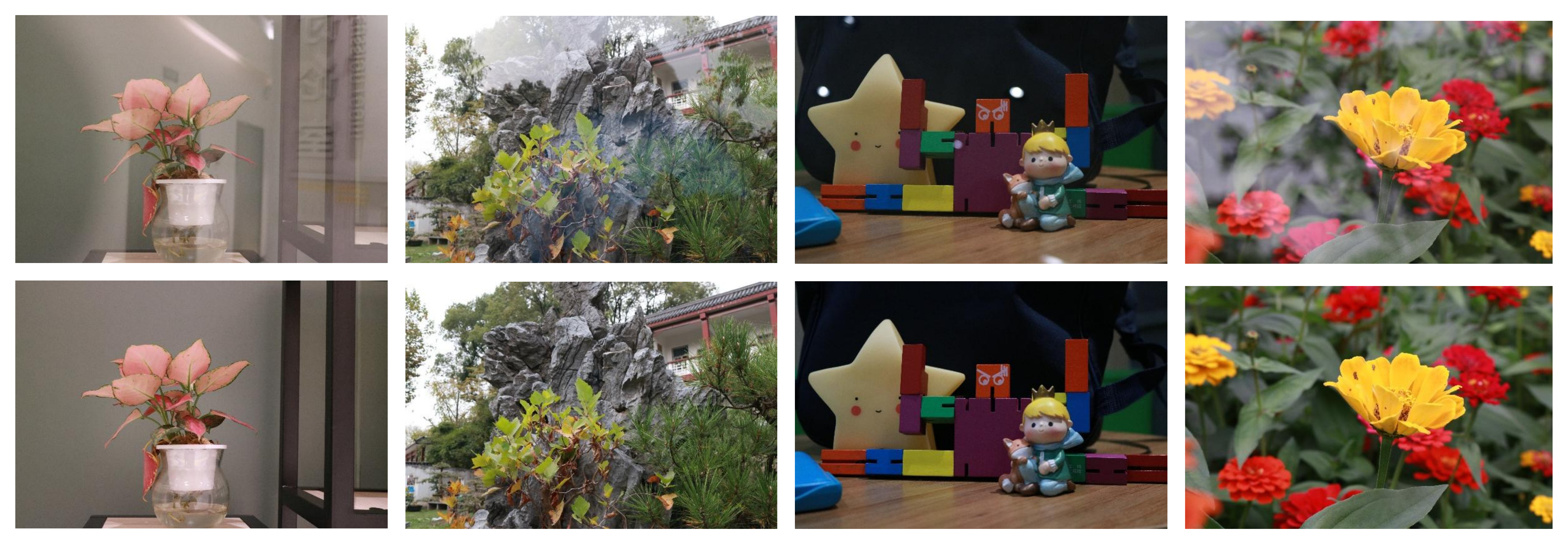}
    \end{center}
    \vspace{-1em}
    \caption{Samples from our real world \textit{Nature} dataset. Top: images with reflection. Bottom: the corresponding ground-truth transmission layers.
    }
    \vspace{-0.5em}
    \label{fig:dataset}
\end{figure}

\begin{figure*}[htbp]
    \begin{center}
    \includegraphics[width=1\textwidth]{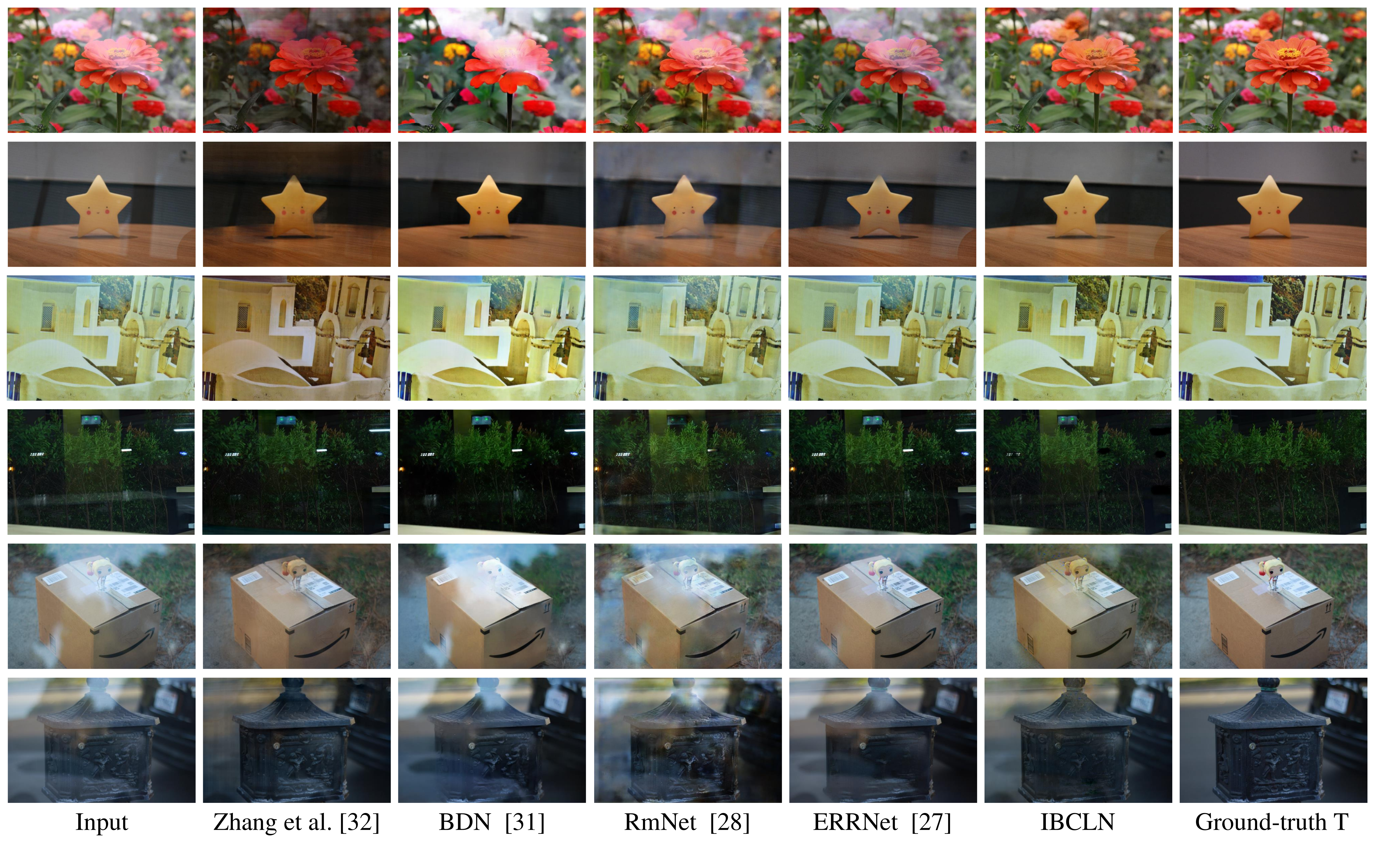}
    \end{center}
    \vspace{-1em}
    \caption{Visual comparison among state-of-the-art approaches and our method on images from three real-world image datasets, namely, \textit{Nature} (Rows 1-2), $SIR^2$ (Rows 3-4) and Zhang et al. (Rows 5-6). More results can be found in the \textit{suppl. material}.
    }
    \vspace{-0.5em}
    \label{fig:method-comparision}
\end{figure*}

Similar to current deep learning methods, our method requires a relatively large amount of data with ground truth for training. Our synthesis model is the same as the recently proposed linear method~\cite{zhang2018single} except for the clipping operation. We utilize their synthetic dataset as well. In our experiments, different methods are evaluated on the publicly available real-world images from the $SIR^2$ datasets~\cite{wan2017benchmarking}, Zhang et al.~\cite{zhang2018single} and the real-world dataset we create. 

Our created dataset, called \texttt{Nature}, contains 220 real-world image pairs: images with reflection and the corresponding ground-truth transmission layers (see samples in Figure \ref{fig:dataset}). We use a Canon EOS 750D for image acquisition. Each ground-truth transmission layer is captured when the portable glass is removed. The dataset is randomly partitioned into a training set and a testing set. We use 200 images for training and 20 images for quantitative evaluation. Inspired by Zhang et al.~\cite{zhang2018single}, we captured the images with the following considerations to simulate diverse imaging conditions: 1) Environments: indoor and outdoor; 2) Lighting conditions: skylight, sunlight, and incandescent; 3) Thickness of the glass slabs: 3 mm and 8 mm; 4) Distance between the glass and the camera: 3 to 15 cm; 5) Camera viewing angles: front view and oblique view; 6) Camera exposure value: 8.0 - 16.0; 7) Camera apertures (affecting the reflection blurriness): f/4.0 — f/16.


\begin{figure*}[h]
    \begin{center}
        \includegraphics[width=1\textwidth]{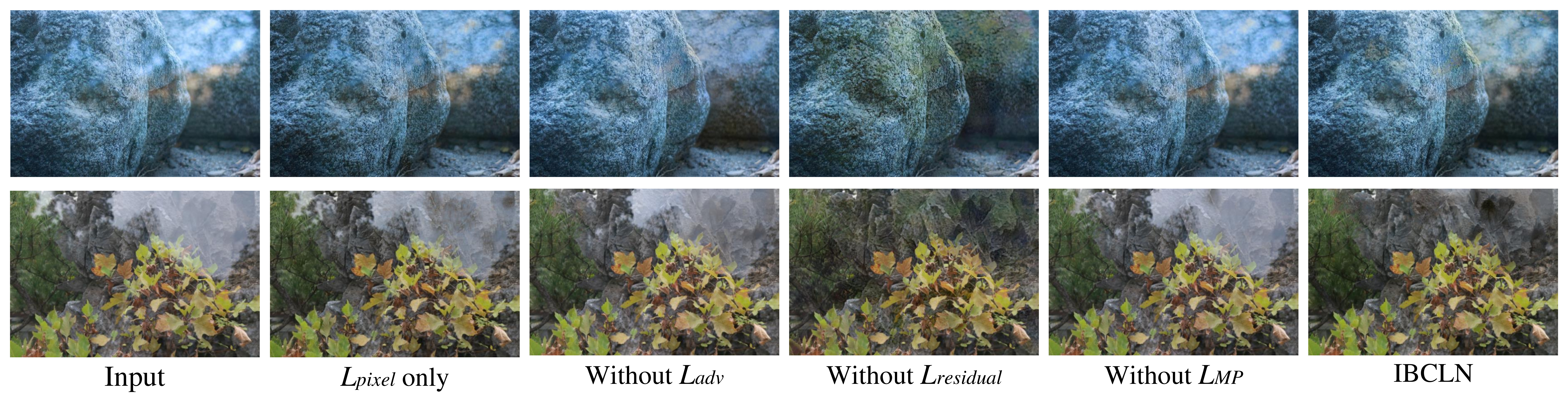}
    \end{center}
    \vspace{-1em}
    \caption{Visual comparison among IBCLN and versions with a modified loss on real-world images. More results are in the \textit{suppl. material}.
    }
    \vspace{-0.8em}
    \label{fig:objective_function}
\end{figure*}

\begin{figure}[h]
    \begin{center}
        \includegraphics[width=0.48\textwidth]{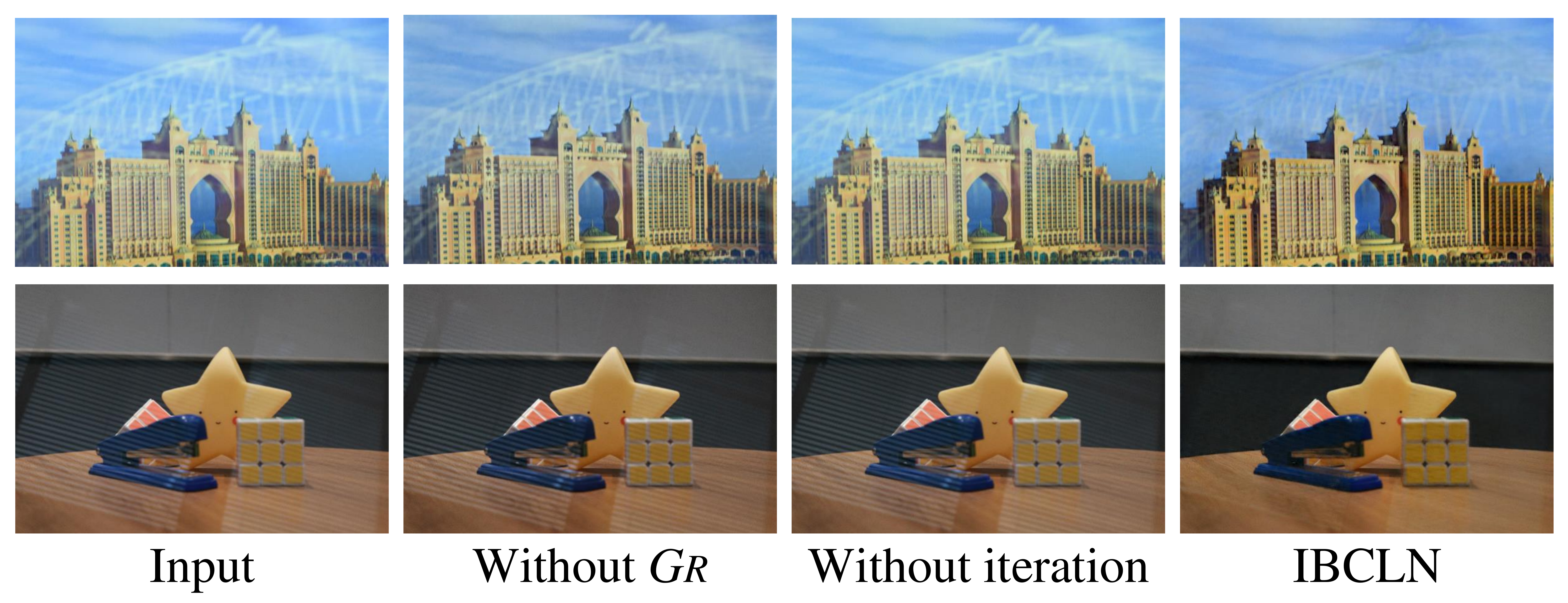}
    \end{center}
    \vspace{-0.8em}
    \caption{Visual comparison among IBCLN and versions with architecture modifications on real-world images. More results can be found in the \textit{suppl. material}. 
    }
    \vspace{-1em}
    \label{fig:architecture}
\end{figure}

\begin{figure}[h]
    \vspace{-1em}
    \begin{center}
        \includegraphics[width=0.48\textwidth]{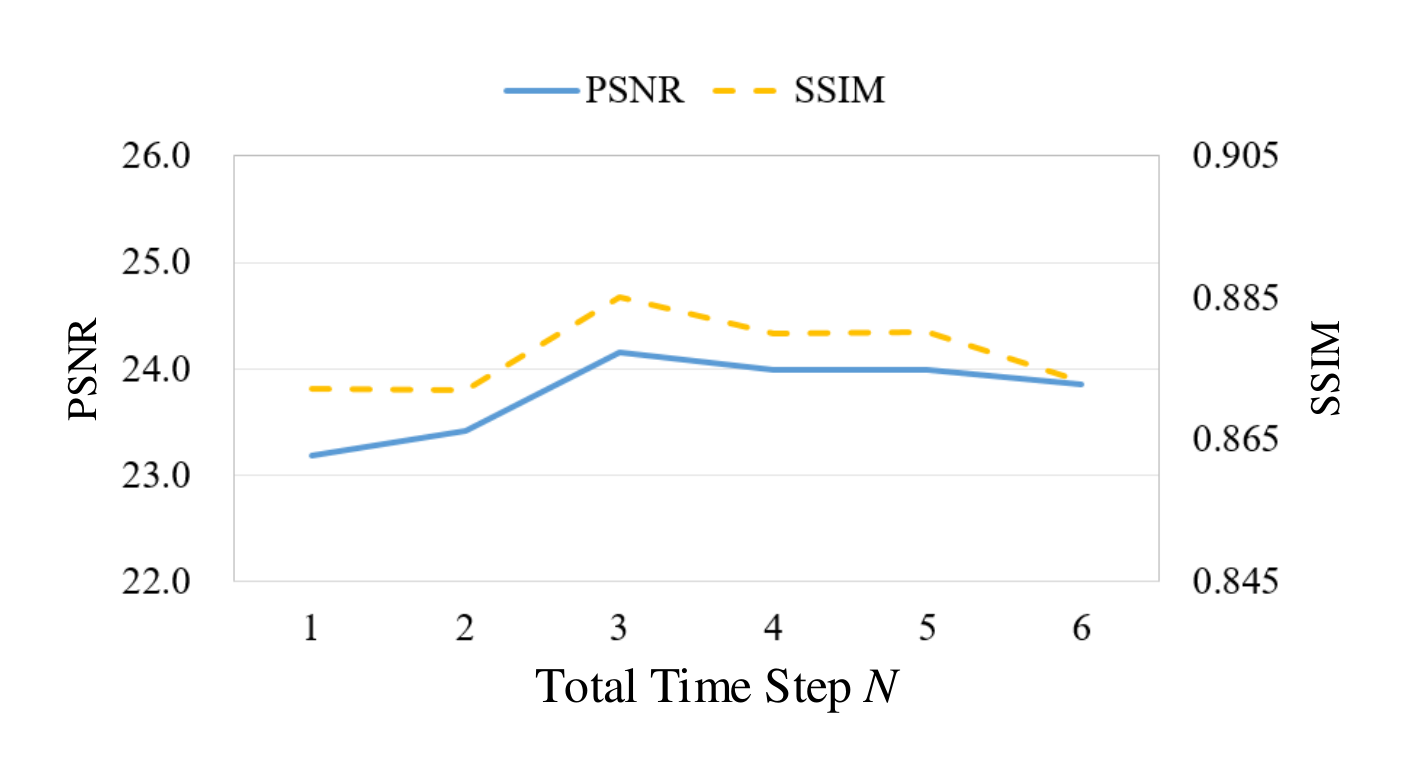}
    \end{center}
    \vspace{-1.2em}
    \caption{Results using different total time steps $N$ in IBCLN on $SIR^2$~\cite{wan2017benchmarking}. Total time steps $N = 3$ yields the best performance.
    }
    \vspace{-0.5em}
    \label{fig:N}
\end{figure}

\vspace{-0.2em}
\subsection{Comparison to State-of-the-art Methods}
\subsubsection{Quantitative Evaluations}
We compare our IBCLN against state-of-the-art methods including CEILNet~\cite{fan2017generic}, Zhang et al.~\cite{zhang2018single}, BDN~\cite{yang2018seeing}, RmNet~\cite{wen2019single} and ERRNet~\cite{wei2019single}. For an apples-to-apples comparison, we finetune each model (if the model provides training code) on our training dataset and report the best result of the original pre-trained model and finetuned version (denoted with a suffix ’-F’). RmNet~\cite{wen2019single} has three models for different reflection types, and we report the best result from among the three models.
\begin{table}[!t]
    \centering
    \caption{Ablation study of IBCLN for architecture on three testing sets. w/o $\bm{G_R}$ means training with only one sub-network $\bm{G_T}$. w/o iteration means the total time steps is 1. Each term contributes to the SIRR performance, and combining all achieves the best results.}
    \footnotesize
    \vspace{2pt}
    \setlength{\tabcolsep}{1.6mm}{
    \begin{tabular}{lcccccc} 
        \toprule
        \multirow{2}{*}{Model}& \multicolumn{2}{c}{Nature} & \multicolumn{2}{c}{Zhang et al.} & \multicolumn{2}{c}{$SIR^2$}\\ \cline{2-7}
         & PSNR & SSIM & PSNR & SSIM & PSNR & SSIM\\ 
         
        \midrule
        w/o $G_R$ & 21.79& 0.759 &20.65 & 0.742&22.36 & 0.868 \\ 
        w/o iteration &  21.82& 0.764  & 20.49 & 0.739& 23.09 & 0.872 \\ 
        Complete &\textbf{23.57}& \textbf{0.783}   &  \textbf{21.86}&  \textbf{0.762} &  \textbf{24.20}&  \textbf{0.884} \\
        \bottomrule
    \end{tabular}}
    \vspace{-1.5em}
    \label{tb:block}
\end{table}

Table \ref{tb:method-com} summarizes results of all the competing methods on five real-world datasets, including three sub-datasets from $SIR^2$~\cite{wan2017benchmarking}, Zhang et al.~\cite{zhang2018single} and our dataset. The number of images in each dataset is shown after the name. The quality metrics include PSNR and SSIM~\cite{wang2004image}. Larger values of PSNR and SSIM indicate better performance. IBCLN achieves the best performance on four of the datasets, but not on 20 images of ``Zhang et al.". As ERRNet~\cite{wei2019single} is developed based on model Zhang et al.~\cite{zhang2018single}, EERNet and Zhang et al. both have better performance on the dataset ``Zhang et al.". In terms of overall performance over all the test datasets, IBCLN surpasses the other methods.

\vspace{-0.5em}
\subsubsection{Qualitative Evaluations}
Figure \ref{fig:method-comparision} presents visual results and the ground truth on real-world images from $SIR^2$~\cite{wan2017benchmarking}, Zhang et al.~\cite{zhang2018single} and our dataset. We select two images from each dataset. It can be seen that Zhang et al.~\cite{zhang2018single} tends to over-remove the reflection layer, while the other baseline methods tend to under-remove. Our model is more accurate and removes most of the undesirable reflections.

\vspace{-0.2em}
\subsection{Controlled Experiments}
For better analyzing our network architecture and the objective function of IBCLN, we separately remove the sub-network $\bm{G_R}$, the iteration step, and the three-loss terms one by one. Then we train new models with the modified networks. The results from these ablations on the architecture are given in Table \ref{tb:block}. The result of a cascade network without LSTM is not shown in the table because it cannot be effectively trained. The ablation study on the loss terms is shown in Table \ref{tb:loss}. And visual comparisons among all the modified networks and IBCLN are displayed in Figure \ref{fig:objective_function} and Figure \ref{fig:architecture}. We observe that using two iterative sub-networks, time steps, $L_{adv}$, $L_{residual}$ and $L_{MP}$ all enhance the performance of IBCLN, and all the blocks and the losses yield different contributions to the removal performance. The complete IBCLN with all structures and objective function terms yields the best results.

To explore how many time steps are appropriate for the predicted transmission to converge, we train the model with different total time steps. Figure \ref{fig:N} exhibits the results. We see that the output approximately converges when total time steps are equal to 3. We experimented with having the network learn the total time steps automatically for different images, but we found that providing this much flexibility causes the performance to decay.
\begin{table}[!t]
    \centering
    \caption{Ablation study of IBCLN for loss terms on three testing sets. Each loss contributes to IBCLN's performance, and combining all achieves the best result.}
    \footnotesize
    \vspace{2pt}
    \setlength{\tabcolsep}{1.6mm}{
    \begin{tabular}{lcccccc} 
        \toprule
        \multirow{2}{*}{Model} & \multicolumn{2}{c}{Nature} & \multicolumn{2}{c}{Zhang et al.} & \multicolumn{2}{c}{$SIR^2$}\\ \cline{2-7}
          & PSNR & SSIM & PSNR & SSIM & PSNR & SSIM\\ 
        \midrule
        $L_{pixel}$ only  & 21.98 & 0.739 & 19.54 & 0.722& 22.91 & 0.843 \\ 
        w/o $L_{adv}$ &23.24& 0.746   &  21.74&  0.755 &  23.86&  \textbf{0.885} \\
        w/o $L_{residual}$ &  22.54&  0.770 &20.98 & 0.755&23.74 & 0.881 \\ 
        w/o $L_{MP}$ &  23.14&  0.744 & 21.47 & 0.734& 22.96 & 0.863 \\ \hline    
        Complete &\textbf{23.57}& \textbf{0.783}   &  \textbf{21.86}&  \textbf{0.762} &  \textbf{24.20}&  0.884 \\
        \bottomrule
    \end{tabular}}
    \vspace{-1.5em}    
    \label{tb:loss}
\end{table}

\section{Conclusion}
We present an Iterative Boost Convolutional LSTM Network (IBCLN) that can effectively remove the reflection from a single image in a cascaded fashion. To formulate an effective cascade network, 
we propose to iteratively refine the transmission and reflection layers at each step in a manner that they can boost prediction quality for each other, and to employ LSTM to facilitate training over multiple cascade steps. The intuition is that a better estimate of the complementary residual reflection can boost the prediction of the transmission, and vice versa. Besides, we incorporate a residual reconstruction loss as further training guidance at each cascade step.  
Moreover, we combine a multi-scale loss with the perceptual loss to form a multi-scale perceptual loss. Quantitative and qualitative evaluations on five datasets (including ours) demonstrate that the proposed IBCLN outperforms state-of-the-art methods on the challenging single image reflection removal problem. 
In future work, we will try our cascaded prediction refinement approach on other image layer decomposition tasks such as raindrop removal, flare removal and dehazing.

\vspace{-0.5em}    
\subsection*{Acknowledgments}
This work is supported by the Fundamental Research Funds for the Central Universities (2019kfyXKJC021) and Microsoft Research Asia.

{\small
    \bibliographystyle{ieee_fullname}
    \bibliography{egbib}
}
    
\end{document}